\documentclass[10pt,twocolumn,letterpaper]{article}

\usepackage{wacv}
\usepackage{times}
\usepackage{epsfig}
\usepackage{graphicx}
\usepackage{amsmath}
\usepackage{amssymb}
\usepackage{booktabs}

\def\wacvPaperID{605} 

\wacvfinalcopy 

\ifwacvfinal
\def\assignedStartPage{9876} 
\fi

\ifwacvfinal
\usepackage[breaklinks=true,bookmarks=false]{hyperref}
\else
\usepackage[pagebackref=true,breaklinks=true,colorlinks,bookmarks=false]{hyperref}
\fi

\ifwacvfinal
\else
\fi

\begin{document}

\title{Modeling dynamic target deformation in camera calibration}

\author{Annika Hagemann\\
Robert Bosch GmbH\\
Computer Vision Research Lab\\ 
{\tt\small annika.hagemann@de.bosch.com}
\and
Moritz Knorr\\
Robert Bosch GmbH\\
Computer Vision Research Lab\\ 
\and
Christoph Stiller\\
Institute of Measurement \& Control Systems\\
Karlsruhe Institute of Technology
}

\maketitle

\begin{abstract}
Most approaches to camera calibration rely on calibration targets of well-known geometry. 
During data acquisition, calibration target and camera system are typically moved w.r.t. each other, to allow image coverage and perspective versatility. We show that moving the target can lead to small temporary deformations of the target, which can introduce significant errors into the calibration result. 
While static inaccuracies of calibration targets have been addressed in previous works, to our knowledge, none of the existing approaches can capture time-varying, dynamic deformations. To achieve high-accuracy calibrations despite moving the target, we propose a way to explicitly model dynamic target deformations in camera calibration.
This is achieved by using a low-dimensional deformation model with only few parameters per image, which can be optimized jointly with target poses and intrinsics.
We demonstrate the effectiveness of modeling dynamic deformations using different calibration targets and show its significance in a structure-from-motion application.
\end{abstract}

\section{Introduction}
\begin{figure}[t!]
	\centering
	\includegraphics[width=0.49\textwidth]{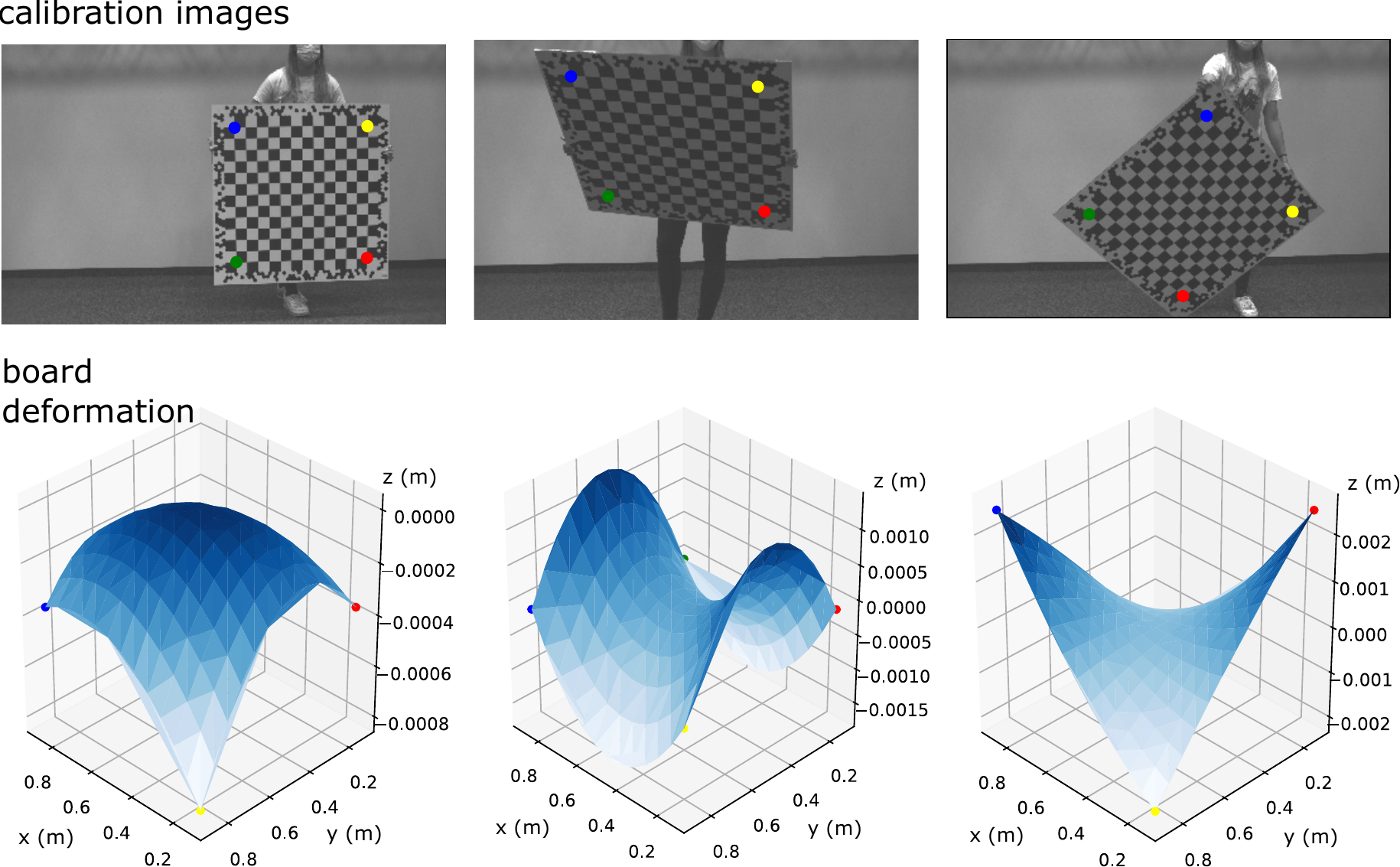}
	\caption{Carrying a calibration target leads to small deformations that can impact the calibration accuracy. The bottom line shows the deformation in each image, estimated using the proposed method. The z-axis is pointing towards the back of the board and the colored dots indicate the corresponding target corners.}
	\label{fig1}
\end{figure}

Target-based camera calibration still constitutes the standard approach to estimating camera intrinsics and extrinsics in 3D computer vision. It relies on the known 3D geometry of precisely printed or measured objects, and can thereby achieve highly accurate estimates for camera characteristics (e.g.~\cite{zhang_flexible_2000,hagemann2021inferring,schops_why_2019,peng_calibration_2019}).
Although target-less methods exist (e.g.~\cite{luong1997self, pollefeys1999self, lopez2019deep}), they have not superseded the classical target-based approaches yet, as the known shape and appearance of calibration targets constitutes a relevant practical advantage.

\par
To obtain an informative set of images for target-based camera calibration, the camera system and target are typically moved w.r.t. each other~\cite{peng_calibration_2019,rojtberg_efficient_2018,richardson_aprilcal:_2013}. This allows capturing images under different angles and distances, leading to a well-constraint calibration problem. Especially when the camera system is rigidly mounted (e.g. on a vehicle, robot or wall), practitioners oftentimes have no choice but to move the target. In fact, many publicly available computer vision datasets show a person carrying the calibration target~\cite{sturm12iros,euroc_paper,usenko18double-sphere}, and computer vision toolboxes are specifically designed for this workflow~\cite{ros, opencv, matlab}.\par
Although carrying calibration targets is common practice, it has, to the best of our knowledge, never been addressed whether, and to which extent, the resulting mechanical forces impact the target geometry and consequently the calibration. While static deformations of a target, e.g. due to printing errors or plastic deformations, have been addressed in various works~\cite{lavest1998we, strobl2008more, albarelli2009robust, strobl2011more, strauss_kalibrierung_2015, strauss2014calibrating, huang2013flexible,xiao2017flexible}, dynamic deformations remained unnoticed.\par
We show that moving or carrying a typical calibration target can lead to small elastic deformations of the material, causing the target to deviate from the assumed geometry (see Fig.~\ref{fig1}). 
Although the magnitude is typically on the order of micrometers to millimeters, we show that these deformations can significantly impact the accuracy of the calibration.\par
To avoid the introduction of these errors into the calibration result, we propose a way to explicitly model the dynamic deformation of calibration targets within the calibration. 
In addition to allowing a static correction of the target geometry, we propose to include a deformation model of low parameter dimensionality, whose parameters are estimated for each individual image.
We demonstrate that this augmentation does not only decrease the calibration error, but also leads to a practically relevant improvement, using the example of egomotion estimation in a Structure-from-Motion (SfM) experiment.

\section{Related work}
Target-based camera calibration is a well-established technique in computer vision and photogrammetry. It relies on the known 3D geometry of precisely printed or measured objects. By associating the known 3D points with the observed image points, camera characteristics are inferred.\par 
Classical photogrammetry uses complex and precisely measured 3D targets that are imaged from a pre-defined set of perspectives~\cite{luhmann2013close}. To avoid this comparatively laborious process, practical computer vision and robotics typically rely on simpler targets and algorithms that do not require a precisely known set of perspectives~\cite{tsai1987versatile, zhang_flexible_2000}. To this end, Zhang et al.~\cite{zhang_flexible_2000} proposed to use planar calibration targets and developed an easy-to-use algorithm for the calibration. Besides the fact that planar targets are comparatively easy to produce, they enable a simple generation of initial values for the target poses using the direct linear transform (DLT)~\cite{hartley_multiple_2004}. As a consequence, the use of planar calibration targets became one of the most widespread approaches to camera calibration~\cite{peng_calibration_2019,matlab,ros,opencv,strauss2014calibrating}. 
However, one critical assumption of Zhang's method, as well as for most other target-based approaches, is the accurate knowledge of 3D target points and the rigidity of the target. This assumption is violated in the case of dynamically deforming targets.\par
Lavest et al.~\cite{lavest1998we} first noticed that small errors in the target geometry can impact the calibration accuracy. They proposed a method for 3D targets that does not only estimate camera intrinsics, but also the 3D geometry of the calibration target. A similar idea was developed for planar calibration targets~\cite{strobl2008more}. Based on the observation that common printers lead to inaccuracies in the printed targets, a 2D-correction was proposed, which could be optimized during calibration~\cite{strobl2008more}. Subsequent work~\cite{albarelli2009robust} proposed to optimize not only a 2D correction, but the full scene structure (3D points) during calibration and reported a significant reduction in residual errors.\par
Strobl and Hirzinger~\cite{strobl2011more} further built upon the idea and proposed a calibration method in which the 3D coordinates of corners are estimated within the bundle adjustment. One of their main contributions was a parametrization that allowed the joint estimation of 3D coordinates and target poses without ambiguities. Their approach showed a significantly improved calibration accuracy when using an inaccurate calibration target (wrinkled paper)~\cite{strobl2011more}. Subsequent works~\cite{huang2013flexible,xiao2017flexible} proposed additional strategies for calibrations with statically imperfect calibration targets. 
\par
While the existing approaches allow for an accurate calibration using an imperfect calibration target, the approaches have focused solely on \emph{static deviations}. This means that only a single set of corrected 3D coordinates is estimated for the entire calibration dataset. To our knowledge, varying, or \emph{dynamic deformations}, that can arise during the calibration process, have not been addressed to date.

\section{Method}

\subsection{Camera projection modeling} \label{sec:camera_projection}
The projection of a 3D world point onto the 2D image can be expressed by a function $\boldsymbol{p}:\mathbb{R}^3\rightarrow\mathbb{R}^2$ that maps a 3D point $\boldsymbol{x}=(x, y, z)^T$ from a world coordinate system to a point $\bar{\boldsymbol{u}}=(\bar{u}, \bar{v})^T$ in the image coordinate system\footnote{We stick to the same calibration formulation and nomenclature as in~\cite{hagemann2021inferring}.}.
This projection can be decomposed into a coordinate transformation from the world coordinate system to the camera coordinate system $\boldsymbol{x} \rightarrow \boldsymbol{x_{c}} =\boldsymbol{R} \boldsymbol{x} + \boldsymbol{t} $ and the subsequent projection from the camera coordinate system to the image coordinate system $\boldsymbol{p_C}:\boldsymbol{x_{c}}\rightarrow \bar{\boldsymbol{u}}$:
\begin{equation}
\bar{\boldsymbol{u}} = \boldsymbol{p}(\boldsymbol{x}, \boldsymbol{\theta}, \boldsymbol{\Pi}) = \boldsymbol{p_C}(\boldsymbol{x_{c}}, \boldsymbol{\theta}) = \boldsymbol{p_C}(\boldsymbol{R} \boldsymbol{x} + \boldsymbol{t}, \boldsymbol{\theta}), 
\end{equation}
where $\boldsymbol{\theta}$ are the \emph{intrinsic} camera parameters and $\boldsymbol{\Pi}$ are the \emph{extrinsic} parameters describing the rotation $\boldsymbol{R}$ and translation $\boldsymbol{t}$ of the camera w.r.t. the world coordinate system.

\subsection{Calibration} \label{sec:calibration_framework}
Without loss of generality, we assume a single chessboard-style calibration target and a single camera in the following. We focus on the monocular case, as it constitutes the harder problem for deformation modeling. For stereo and multi-camera calibration, where the target is visible in more than one camera, the problem becomes more constraint. We argue that, if deformation modeling is successful for monocular calibration, it can be easily extended to the multi-camera case.\par
The calibration target contains a set of $N_\mathcal{C}$ easily detectable 3D points (e.g. chessboard corners $\mathcal{C}=\{\text{corner}_i\}_{i=1}^{N_\mathcal{C}}$ for checkerboard targets). The 3D coordinates of each target point $i$ in the world (target) coordinate system are assumed to be known as $\boldsymbol{x}_{i}=(x_i, y_i, z_i)^T$. 
For simplicity and without loss of generality we assume the planar checkerboard surface to coincide with the world's x-y-plane, such that $\boldsymbol{x}_{i}=(n_i d, m_i d, 0)^T$ where $n_i$ and $m_i$ are the index of corner $i$ and $d$ is the grid width of the checkerboard. \par
The calibration dataset is a set of $N_\mathcal{F}$ images $\mathcal{F}=\{\text{frame}_j\}_{j=1}^{N_\mathcal{F}}$, where the calibration target is visible from different perspectives. The image coordinates of chessboard-corners in each image are determined by a corner detection algorithm~\cite{strauss2014calibrating}, giving the observations $\boldsymbol{u}_{ij}=(u_{ij}, v_{ij})^T$ for each corner $i$ in each image $j$. 
As corner detection is not perfectly accurate, the observed coordinates $\boldsymbol{u}_{ij}$ will deviate from the true image points $\boldsymbol{\bar{u}}_{ij}$ by a certain error $\boldsymbol{\epsilon_d}$, called observational noise. This error is typically assumed to be independent identically distributed (i.i.d.) $\boldsymbol{\epsilon_d} \sim \mathcal{N}(\boldsymbol{0}, \sigma_d^2 \boldsymbol{I})$ where $\sigma_d^2$ is called the detector variance. \par
Following~\cite{zhang_flexible_2000}, camera intrinsics $\boldsymbol{\theta}$ and target poses $\{\boldsymbol{\Pi_j}\}_{j=1}^{N_\mathcal{F}}$ are estimated by minimizing a calibration cost function, defined by the sum of squares of reprojection errors
\begin{equation}\label{eq:standard}
\epsilon_{\mathrm{res}}^2 = \sum_{j\in \mathcal{F}} \sum_{i \in \mathcal{C}} ||\boldsymbol{u}_{ij} - \boldsymbol{p}(\boldsymbol{x}_{i}, \boldsymbol{\theta}, \boldsymbol{\Pi}_j)||^2.
\end{equation}
Having generated initial values for the target poses using the DLT, the calibration cost is optimized using a non-linear least-squares algorithm, giving parameter estimates
$(\boldsymbol{\hat{\theta}}, \{\boldsymbol{\hat{\Pi}_j}\}_{j=1}^{N_\mathcal{F}}) = \text{argmin}(\epsilon_{\mathrm{res}}^2)$.\par 
For simplicity, all formulas are written for the case where the target is fully visible in the image, although the methods are not restricted to this case. Furthermore, we present formulas for non-robust optimization. To reduce the impact of potential outliers, robustification (e.g. a cauchy kernel) can be included. 

\subsection{Modeling static board deformation}
Static board deformations can be modeled by introducing an additional set of parameters $\{\boldsymbol{\Delta x_i}\}_{i\in C}$ that describe the deviation of the 3D coordinate of the corner from the assumed planar grid~\cite{strobl2011more}. Importantly, 7 degrees of freedom must remain fixed to prevent ambiguity with the target poses\footnote{Six degrees of freedom corresponding to the target pose, plus one to fix the absolute scale~\cite{strobl2011more}.}. In~\cite{strobl2011more} this is achieved by setting the correction of two corners to zero, $\boldsymbol{\Delta x_1}=(0, 0, 0)$, $\boldsymbol{\Delta x_2}=(0, 0, 0)$ and furthermore fixing the last coordinate of a third corner $\boldsymbol{\Delta x_3}=(\Delta x_3, \Delta y_3, 0)$. Here the choice of the fixed corners is free, as long as they are non-collinear~\cite{strobl2011more}.
The cost function with global deformation is then given by
\begin{equation}\label{eq:static}
\epsilon_{\mathrm{res}}^2 = \sum_{j\in \mathcal{F}} \sum_{i \in \mathcal{C}} ||\boldsymbol{u}_{ij} - \boldsymbol{p}(\boldsymbol{x}_{i}, \boldsymbol{\theta}, \boldsymbol{\Pi}_j, \boldsymbol{\Delta x_i})||^2.
\end{equation}

\begin{figure*}[t!]
	\centering
	\includegraphics[width=0.99\textwidth]{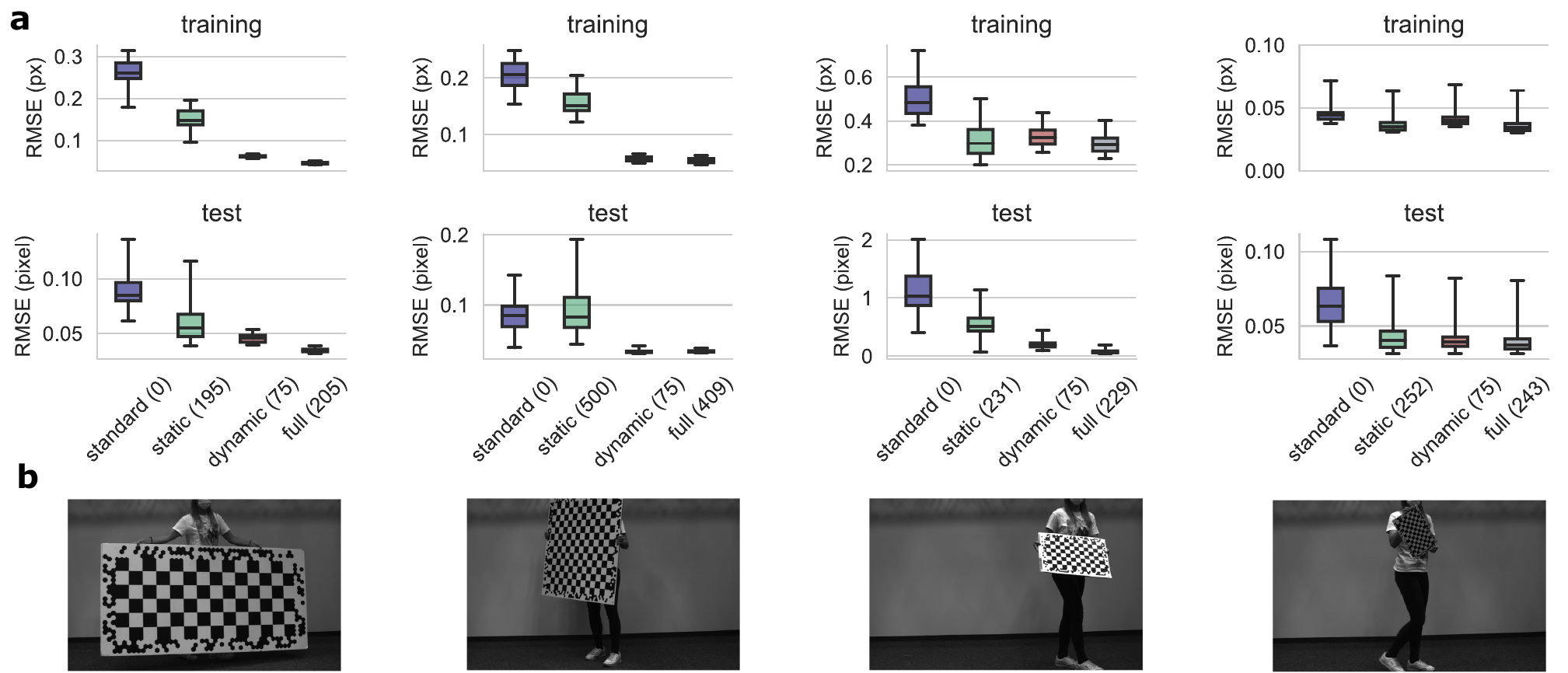}
	\caption{Comparison of the different calibration methods for different types of calibration targets. \textbf{a} Training error and test error of the different calibration methods (\emph{standard}, \emph{static}, \emph{dynamic} and \emph{full}) on datasets with different targets. The number in brackets indicates the number of deformation parameters. \textbf{b} Example images of the targets T1 (1~m $\times$ 2~m $\times$ 4~cm, styrofoam composite), T2 (1~m $\times$ 1~m $\times$ 0.6~cm, aluminum polyethylene composite), T3 (60~cm $\times$ 42~cm $\times$ 0.7~cm, paper glued to cardboard) and T4 (37.5~cm $\times$ 25~cm $\times$ 0.25~cm, aluminum polyethylene composite). We used an Allied Vision camera MG-235B with XENOPLAN 1.4/17 lens. The results and images correspond column-wise. The results for the reference dataset, where the target is lying statically on the ground, are shown in the Supplementary Fig.~S1.}
	\label{fig3}
\end{figure*}

\subsection{Modeling dynamic board deformation}
Modeling dynamic board deformations requires a time-dependent correction $\boldsymbol{\Delta x_{i}}(t)$ over time $t$. As the calibration dataset consists of a discrete set of images, time-dependence can be implemented by allowing for a different correction $\boldsymbol{\Delta x_{ij}}$ in every image $j$. For the calibration problem to still be well-conditioned, the correction cannot be modeled freely for every 3D point, as in equation (\ref{eq:static}). Instead, a low-dimensional parameterization with few parameters per image is needed.\par
We make use of the fact that physical deformations lead to spatially correlated deviations. If deviations $\boldsymbol{\Delta x_{ij}}$ of different corners $i$ are correlated, they can be approximated with a function that only depends on the original corner coordinates $\boldsymbol{x_i}$ and an additional set of parameters $\boldsymbol{\beta}_j$ that can differ in each image $j$:
\begin{equation}
\boldsymbol{\Delta x_{ij}} = \boldsymbol{f(x_i, \beta_j)}
\end{equation}
The parameters $\{\boldsymbol{\beta_j}\}_{j=1}^{N_\mathcal{F}}$ can be estimated jointly with intrinsics $\boldsymbol{\theta}$ and target poses $\{\boldsymbol{\Pi_j}\}_{j=1}^{N_\mathcal{F}}$ during calibration.
The calibration cost function is then given by
\begin{equation}\label{eq:dynamic}
\epsilon_{\mathrm{res}}^2 = \sum_{j\in \mathcal{F}} \sum_{i \in \mathcal{C}} ||\boldsymbol{u}_{ij} - \boldsymbol{p}(\boldsymbol{x}_{i}, \boldsymbol{\theta}, \boldsymbol{\Pi}_j, \boldsymbol{\beta_j})||^2.
\end{equation}
While this general formulation can be applied to any target, the specific choice of the function $\boldsymbol{f(x_i, \beta_j)}$ may depend on the target's geometry and material. For planar checkerboard targets with comparatively rigid/stiff material (e.g. aluminum polyethylene composite), we propose a simple paraboloid to model the major deformations in the z-coordinate of the target
\begin{equation}
\boldsymbol{f(x_i, \beta_j)} = (0, 0, a_jx_{i}^2+b_jy_{i}^2+c_jx_{i}y_{i}),
\end{equation}
with $a_j, b_j, c_j \in \mathbb{R}$. Here  $x_i$ and $y_i$ are the corner coordinates in the board coordinate system, the origin of which we assume to be in the center of the board. The formulation can be extended with additional terms, as long as the terms do not lead to an ambiguity with the target poses. A constant term (0th order), for instance, describes a translation in z-direction which would result in an ambiguity with the target translation. A linear term (first order), could result in an ambiguity with the target rotation.\par
We chose this simple model to demonstrate that no complex modeling of mechanical forces and deformations is needed to obtain a significant improvement in calibration accuracy. Clearly, more sophisticated models that preserve area and angles on the board, or that account for mechanical properties of the material could also be used. However, our experiments show that even this simple model leads to a significant reduction of calibration errors, which suggests that the model is capable of capturing the major deformations of a typical planar target.

\subsection{Combining a static in-plane with a dynamic out-of-plane deformation}
In practice, a superposition of static and dynamic deformations is likely. A static deformation can be caused by an imprecise print, while dynamic deformations can arise from gravity induced sag while recording calibration images. To cover both, we propose to superimpose a static in-plane deformation in the $x$-$y$-plane with dynamic out-of-plane deformations in the target's $z$-direction.
When including a static in-plane deformation $(\Delta x_{i}, \Delta y_{i})$, we have to constrain four degrees of freedom\footnote{Two degrees of freedom for the target translation in x- and y-direction, one for the rotation within the $z=0$ plane and one to fix the absolute scale.}, which can be achieved by setting 
$(\Delta x_{1}, \Delta y_{1})=(0, 0)$ (fixes the target position) and $(\Delta x_{2}, \Delta y_{2})=(0, 0)$ (fixes the target scale and orientation within the $z=0$ plane). The cost function is then given by
\begin{equation}\label{eq:full}
\epsilon_{\mathrm{res}}^2 = \sum_{j\in \mathcal{F}} \sum_{i \in \mathcal{C}} ||\boldsymbol{u}_{ij} - \boldsymbol{p}(\boldsymbol{x}_{i}, \boldsymbol{\theta}, \boldsymbol{\Pi}_j, \boldsymbol{\beta_j}, \Delta x_i, \Delta y_i)||^2.
\end{equation}

\subsection{Extension to multi-camera calibration}
Modeling board deformations is even better constraint in the multi-camera case, as compared to the monocular case. Therefore, the methods that we proposed for monocular calibration are directly applicable to stereo- and multi-camera calibration.
For a set of $N_\mathcal{K}$ cameras $\mathcal{K}=\{\text{camera}_k\}_{k=1}^{N_\mathcal{K}}$, the cost function for modeling dynamic deformations is given by
\begin{equation}
\epsilon_{\mathrm{res}}^2 = \sum_{k\in \mathcal{K}} \sum_{j\in \mathcal{F}} \sum_{i \in \mathcal{C}} ||\boldsymbol{u}_{ijk} - \boldsymbol{p}(\boldsymbol{x}_{i}, \boldsymbol{\theta}_k, \boldsymbol{\pi}_k, \boldsymbol{\Pi}_j, \boldsymbol{\beta_j})||^2,
\end{equation}
where $\boldsymbol{\pi}_k$ with $k=1,...,N_\mathcal{K}$ denotes the relative poses of all cameras w.r.t. the first camera, and $\boldsymbol{\Pi}_j$ is the pose of the calibration target w.r.t. the first camera in image $j$.

\section{Experimental setup}
To investigate the effect of target deformations on the calibration, we collected multiple calibration datasets and compared four different calibration methods:
\begin{itemize}
	\item \textbf{standard}: the calibration target is assumed to be planar with fixed board coordinates $\boldsymbol{x}_{i}=(n_i d, m_i d, 0)^T$. This is the standard approach in camera calibration (equation (\ref{eq:standard})).
	\item \textbf{static}: the target is assumed to statically deviate from the assumed geometry. One global correction is estimated for the full dataset (equation (\ref{eq:static})).
	\item \textbf{dynamic (ours)}: the target is assumed to deform dynamically (different in each image), where the deformation is described by a paraboloid (equation (\ref{eq:dynamic})).
	\item \textbf{full (ours)}: a static in-plane deformation is combined with a dynamic out-of-plane deformation (equation (\ref{eq:full})).
\end{itemize}

\subsection{Calibration using different targets}
We used four calibration targets of different size and material (see Fig.~\ref{fig3}\textbf{b}). All targets are coded to enable a clear association of corners~\cite{strauss2014calibrating}. Targets T1, T2 and T4 are high-quality, precisely manufactured targets, while target T3 is self-made by printing the target on two sheets of paper and fixing them on cardboard. This should mimic a quick and comparatively inaccurate attempt. To investigate the effect of dynamic board deformations, we collected datasets in which the targets were carried by a person and moved w.r.t. the camera (see Fig.~\ref{fig3}\textbf{b}). The full calibration dataset for each target contained more than 500 images. Out of each of these datasets, we drew 50 random subsets containing 25 images each, on which we ran a calibration with each method. This allowed a statistical analysis of the different calibration methods. 
\par
As a reference calibration, we collected a dataset of 650 images with the calibration target T2 lying statically on the ground, while moving the camera. In this setup, dynamic deformations can be ruled out and only a static correction of the board may be needed. We therefore used the \emph{static} method and all 650 images to obtain a reference calibration. To verify that the calibration did not change during recording, we collected a second reference dataset after performing all recordings (see Fig.~\ref{fig_dist}).\par
In all calibrations, we used a pinhole model with radial distortion
\begin{align}
u =f_x \frac{x_{c}}{z_{c}} (1 + k_1 r^2 + k_2  r^4 + k_3  r^6) + \mathrm{ppx},\nonumber\\ 
v = f_y \frac{y_{c}}{z_{c}} (1 + k_1 r^2 + k_2  r^4 + k_3  r^6) + \mathrm{ppy},
\end{align}
where $ r = \sqrt{\left(\frac{x_{c}}{z_{c}}\right)^2+\left(\frac{y_{c}}{z_{c}}\right)^2}$, $f_x$ and $f_y$ are the focal lengths, $(\mathrm{ppx}, \mathrm{ppy})$ is the principal point and $k_1$, $k_2$, $k_3$ are the distortion parameters.

\subsection{Computing the training and test error}
To assess the accuracy of calibration results obtained with different methods and targets, we made use of different metrics. The most straight-forward approach is the inspection of calibration residuals and the computation of the root mean squared error (RMSE). However, this error corresponds to a \emph{training} error, as it is computed on the dataset on which the parameters were optimized. Therefore, it does not capture a possible overfitting, which could occur when introducing additional parameters.
To capture overfitting, we additionally computed a \emph{test} error on an independent test dataset. For that purpose, we used the reference calibration dataset, where the calibration target lied statically on the ground. To compute the test error for a given set of estimated intrinsics $\boldsymbol{\hat{\theta}}$, we then ran a "reduced" calibration on the reference dataset. 
In this reduced calibration, the intrinsics $\boldsymbol{\hat{\theta}}$ were fixed, and only the target poses were optimized\footnote{To correct the static imperfection of the target, we included a fixed, static correction $\{\boldsymbol{\Delta x_i}\}_{i\in C}$ that had previously been estimated by the reference calibration.} (see also~\cite{richardson_aprilcal:_2013}). The residual errors of this reduced calibration are a test error for the accuracy of estimated intrinsics.

\subsection{Computing the mapping error}
An alternative approach to assess the accuracy of the different intrinsic calibration results is the direct comparison with the ground-truth intrinsic parameters. As there is no ground-truth available in real-world experiments, we used the reference calibration as an approximation for the ground-truth.\par 
To compare the different sets of intrinsics, we used \emph{mapping error} in image space~\cite{beck_generalized_2018, hagemann2020bias, hagemann2021inferring}. This metric propagates the differences in the individual parameters to a difference in image space. It is obtained by projecting a regular grid of image points from the image to 3D points using one set of intrinsics, and then projecting these 3D points back to the image using the other set of intrinsics. The mapping error is then defined by the RMSE of original image points compared to the projected image points.

\subsection{Structure-from-Motion}
To investigate the practical impact of dynamic board deformations, we use a Structure-from-Motion experiment. We collected a dataset with the camera moving around a building and returning to the starting point. To make sure that start and endpoint are exactly identical, we added the first image again to the end of the dataset. \par
We used AliceVision meshroom~\cite{meshroom}, an open-source 3D reconstruction software, to infer the camera trajectory and 3D structure of the building. Importantly, we only allowed for pairwise sequential feature matching, and deactivated the automatic loop detection. Thereby, calibration errors can accumulate along the trajectory.\par The accuracy of the trajectory can now be assessed by the loop closure error. As in~\cite{peng_calibration_2019}, we assess the loop closure error in image space: all 3D points that were reconstructed based on features extracted in the \emph{first} image are projected to the image using the estimated pose of the \emph{last} image. The RMSE of the original feature points compared to the projected 3D points then serves as a measure for the loop-closure error. The main advantage of this assessment is that it is unaffected by the scale ambiguity of monocular SfM.

\section{Results}
\begin{figure*}[ht!]
	\centering
	\includegraphics[width=0.99\textwidth]{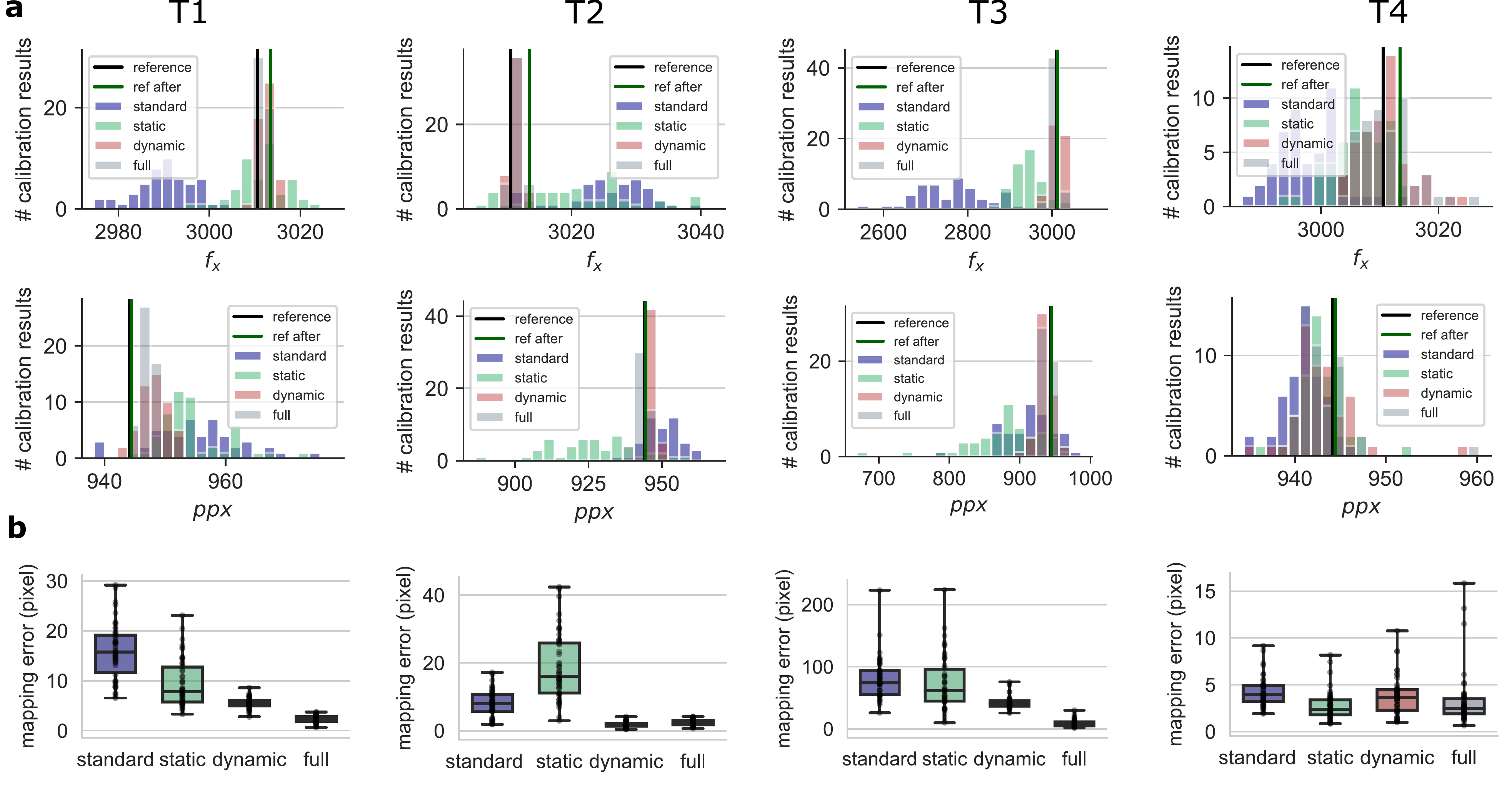}
	\caption{\textbf{a} Distribution of estimated intrinsic parameters using the different targets (T1, T2, T3, T4) and calibration methods (standard, static, flexible, full). The focal length $f_x$ and the principal point ppx are depicted. For all other parameters see Supplementary Figs.~S2 and S3. The vertical lines show the reference calibrations before and after all experiments. \textbf{b} Error in estimated intrinsic parameters w.r.t. the reference calibration, measured in terms of the mapping error in image space. High values indicate a large deviation from the reference calibration.}
	\label{fig_dist}
\end{figure*}
\subsection{Reprojection errors on training and test dataset}

To analyze the impact of board deformations, we ran 50 calibrations with each of the four calibration methods \emph{standard}, \emph{static}, \emph{dynamic} and \emph{full}, for each of the calibration targets shown in Fig.~\ref{fig3}b. The reprojection errors are shown in Fig.~\ref{fig3}a. For all targets, the RMSE was reduced by modeling board deformation. 
Targets T1 and T2 show similar results with $\mathrm{RMSE}_\mathrm{standard}>\mathrm{RMSE}_\mathrm{static}>\mathrm{RMSE}_\mathrm{dynamic}>\mathrm{RMSE}_\mathrm{full}$. Interestingly, this order does not reflect the number of additional parameters: although the \emph{static} method contains more additional parameters than the \emph{dynamic} method, it leads to a smaller reduction in the RMSE. Target T3 shows the same trend, but higher RMSE values, as it is a self-made, less accurate target. Target T4, on the other hand, shows comparatively low RMSE values because (i) it is small and light such that less deformations are to be expected and (ii) due to its size it only covers small image regions, so that errors could partially be compensated by adjusting the pose estimate. For this target, deformation modeling only had a small effect on the RMSE. In summary, modeling dynamic board deformation led to a clear reduction in reprojection errors in three out of the four targets.\par
The reprojection error is always a \emph{training error}. As the number of parameters increases when modeling board deformations, a reduction in the RMSE is to be expected regardless of whether they lead to an improvement of the calibration. In particular, adding a large number of parameters comes with the risk of overfitting to the given dataset. We therefore additionally assessed the \emph{test error}, using the reference dataset with the target lying statically on the ground.\par
Across most calibrations, the test error was highly correlated with the training error (see Fig.~\ref{fig3}a). This suggests that the reduction in training errors was caused by an improved calibration rather than overfitting. For the \emph{static} method, however, the test error for target T2 was increased. This suggests that the introduction of a static deformation led to less accurate estimates of intrinsic parameters. One possible explanation is that target T2 showed dynamic deformations which the \emph{static} method could not model. Estimating a fixed correction when deformations are in fact dynamic resulted in wrong intrinsics. In summary, Fig.~\ref{fig3} shows that for the three targets with larger physical extent, modeling dynamic deformations (\emph{dynamic} or \emph{full}) led to a significant reduction in the test error.

\subsection{Intrinsic parameters and mapping error}

We compared the intrinsic parameters obtained with each of the four calibration methods \emph{standard}, \emph{static}, \emph{dynamic} and \emph{full}.
Fig.~\ref{fig_dist}\textbf{a} shows the distributions of the estimated focal length $f_x$ and the principal point ppx for all targets and methods. Additionally, we annotated the reference calibrations, obtained before and after the different recordings. 
All parameters show statistical fluctuations, i.e. the results vary depending on the image sample. However, the variance tended to be higher for the methods \emph{standard} and \emph{static}, as compared to \emph{dynamic} and \emph{full}. In fact, using dynamic deformation modeling, the parameters remained comparatively close to the reference calibration across all image samples. \par
To express the deviation in the individual parameters as a single number, we computed the mapping error w.r.t. the reference calibration~\cite{hagemann2020bias, beck_generalized_2018, hagemann2021inferring}. Fig.~\ref{fig4}\textbf{b} shows the resulting values for all targets and methods. Consistent with the test error (Fig.~\ref{fig3}), the mapping error for targets T1, T2 and T3 reduced significantly when dynamic deformations were modeled (\emph{dynamic} or \emph{full}). Only for target T4, the choice of the calibration method did not lead to a significant difference in calibration results. \par

\section{Structure-from-Motion}
\begin{figure}[ht!]
	\centering
	\includegraphics[width=0.47\textwidth]{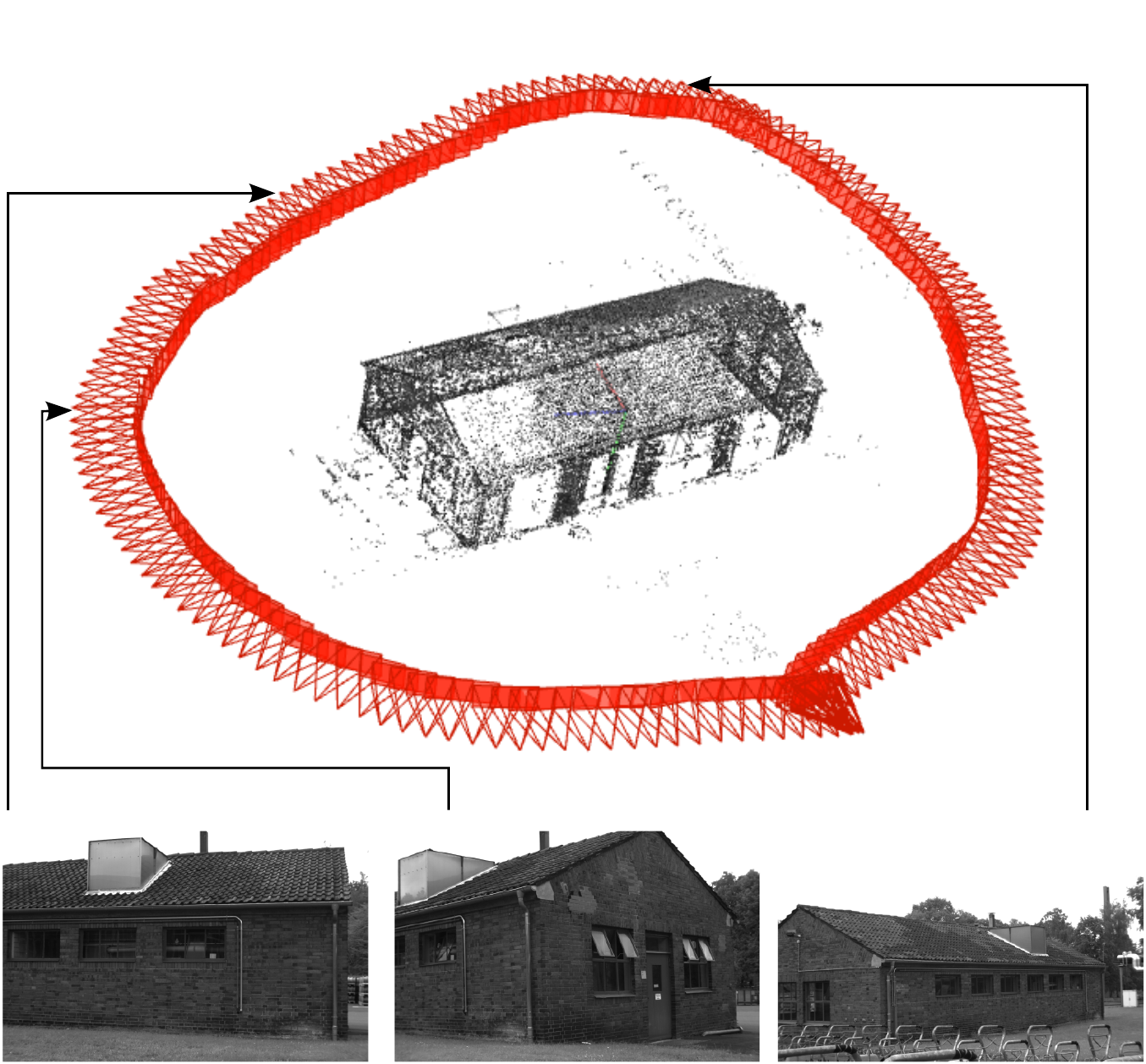}
	\caption{Structure-from-Motion experiment. The camera was moved around a building and then returned to the starting point. The bottom row shows exemplary images. }
	\label{fig4}
\end{figure}
\begin{figure}[h!]
	\centering
	\includegraphics[width=0.49\textwidth]{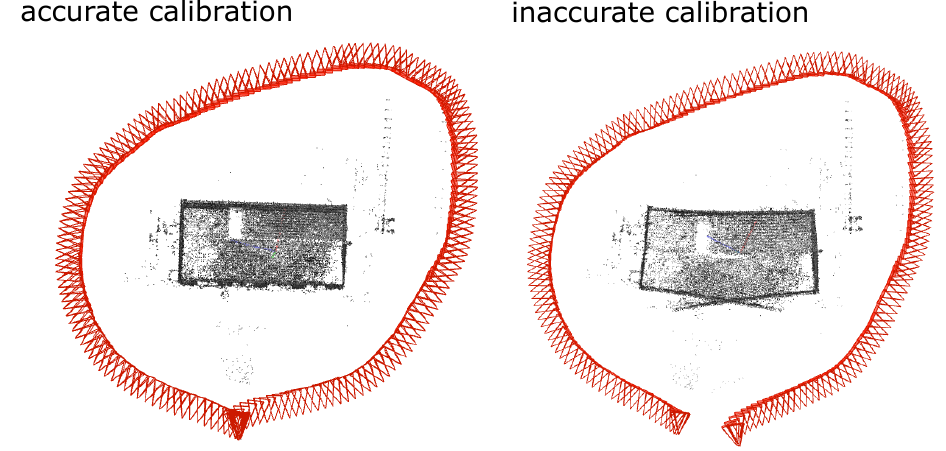}
	\caption{Top-view of two exemplary reconstruction results. Using an inaccurate calibration leads to significant errors in the 3D reconstruction, as well as in the estimated trajectory.}
	\label{fig5}
\end{figure}
To investigate the practical impact of dynamic board deformations, we used a Structure-from-Motion experiment. Using the same camera as before, we collected a dataset where the camera is moved in a loop around a building. Fig.~\ref{fig4} shows the reconstructed scene and the camera trajectory, as well as some example images (for visualization we used COLMAP~\cite{schoenberger2016sfm,schoenberger2016mvs} since we preferred the visualization style; quantitative analyses were performed with AliceVision meshroom~\cite{meshroom}). The reconstruction shown in Fig.~\ref{fig4} is based on the reference calibration. If the assumed calibration is accurate, and sufficient features could be extracted and matched, the SfM algorithm is expected to return an accurate reconstruction. If, however, the calibration used within the SfM is highly inaccurate, less accurate reconstructions and trajectory estimates are expected. \par
Fig.~\ref{fig5} shows a top-view on the reconstruction using the reference calibration, and an inaccurate calibration (obtained from target T3 using the standard method). As loop closure detection is disabled, the errors can accumulate along the trajectory. For the inaccurate calibration, this results in a large pose gap between first and last image. Furthermore, the 3D structure of the building contains substantial errors. This shows that the accuracy of a calibration has a clear practical impact.\par
To assess the different calibration methods in this practical experiment, we ran the SfM experiment using the intrinsics obtained from the different calibration runs (using different targets and deformation models), and computed the loop closure error in image space. Table~\ref{table:1} shows the results. Across all targets, modeling dynamic board deformations led to significantly lower loop closure errors. This demonstrates that dynamic board deformations are not a minor effect, but that they can have a practically relevant impact in applications.
\begin{table}[ht!]
	\centering
	\begin{tabular}{ lll } \toprule
		target & calibration method & loop closure RMSE (px) \\ \midrule
		T1 &  standard & $16.1 \pm 5.7 $  \\ 
		T1 &  static & $5.1 \pm 3.7$  \\ 
		T1  & dynamic (ours) & $3.5 \pm 1.8$  \\ 
		T1  & \textbf{full (ours)} & \textbf{1.5}~$\pm$~\textbf{0.7} \\ \midrule
		T2  & standard & $9.2 \pm 5.0$  \\ 
		T2 &  static & $11.8 \pm 7.6$ \\ 
		T2 &  \textbf{dynamic (ours)}  &  \textbf{1.4}~$\pm$~\textbf{0.6}  \\ 
		T2 &  full (ours) & $1.8 \pm 0.6$  \\ \midrule
		T3 &  standard & $248.7 \pm 109.3$  \\ 
		T3 &  static & $49.8 \pm 31.5$  \\ 
		T3 &  dynamic (ours) & $14.2 \pm 9.9$  \\ 
		T3 & \textbf{full (ours)} &\textbf{8.0}~$\pm$~\textbf{5.6} \\ \midrule
		T4 &  standard &$9.7 \pm 3.2$  \\ 
		T4  & static & $5.1 \pm 2.8$  \\ 
		T4  & \textbf{dynamic (ours)} & \textbf{3.6}~$\pm$~\textbf{2.4}  \\ 
		T4  & \textbf{full (ours)} & \textbf{3.6}~$\pm$~\textbf{2.4}  \\ 
		\bottomrule
	\end{tabular}
	\vspace{2mm}
	\caption{Reprojection error (root mean squared error, RMSE) in the Structure-from-Motion experiment. The error reflects the loop closure error of the trajectory in image space. The values are mean and standard deviation across 50 calibrations.}
	\label{table:1}
\end{table}

\section{Discussion}
We have shown that carrying a calibration target can lead to dynamic deformations that impact the accuracy of calibration results. As moving a calibration target in front of the camera constitutes one of the most widespread approaches to camera calibration, we consider it important for practitioners to be aware of this effect.\par
We proposed a way to explicitly model dynamic target deformations in the calibration. We thereby extended the existing work on static board deformation that only accounted for a single global correction across the entire dataset. Our results suggest that incorporating dynamic deformations leads to significantly more accurate and precise calibration results.\par
Clearly, the magnitude of the effect depends on the target's size, shape, weight and material. Our results suggest that deformations are more relevant for large calibration targets, while the effect was smaller for the small, light target. Furthermore, self-made targets clearly show more deformations than professional targets. However, even for high-quality targets, the incorporation of target deformations led to more accurate calibration results.\par
In general, modeling static or dynamic board deformations introduces additional parameters into the calibration. Depending on the amount of data, this can come with a risk of overfitting to the data, as well as a risk of ambiguities (e.g. wrongfully attributing lens distortions to a deformed target). However, we did not observe such an effect in our experiments. Instead, the calibration results showed a lower variance than the standard calibration results. 
To ensure that deformation modeling did not lead to overfitting on a given dataset, one can either evaluate the calibration result on a separate test dataset, or evaluate the model uncertainty, e.g. using the expected mapping error~\cite{hagemann2021inferring}. Furthermore, model selection criteria, such as the geometric akaike information criterion~\cite{kanatani1998geometric} and the bias ratio~\cite{hagemann2021inferring} can be applied to decide whether deformation modeling is needed for a specific dataset.
\par
Importantly, our results suggest that a simple paraboloid is sufficient to achieve a significant improvement of calibration results. Due to this simplicity, the proposed model can be easily incorporated into existing calibration frameworks. In future, one could also account for temporal correlations in deformations, to further reduce the number of additional parameters. Furthermore, more elaborate deformation models could be tried, which explicitly account for the mechanical properties of the target.

{\small
	\bibliographystyle{ieee_fullname}
	\bibliography{egbib}
}

\newpage

\newcommand{\beginsupplement}{%
	\setcounter{table}{0}
	\renewcommand{\thetable}{S\arabic{table}}%
	\setcounter{figure}{0}
	\renewcommand{\thefigure}{S\arabic{figure}}%
	\setcounter{section}{0}
}

\onecolumn
\section*{Supplementary material}
\beginsupplement

\begin{figure*}[h!]
	\centering
	\includegraphics[width=0.89\textwidth]{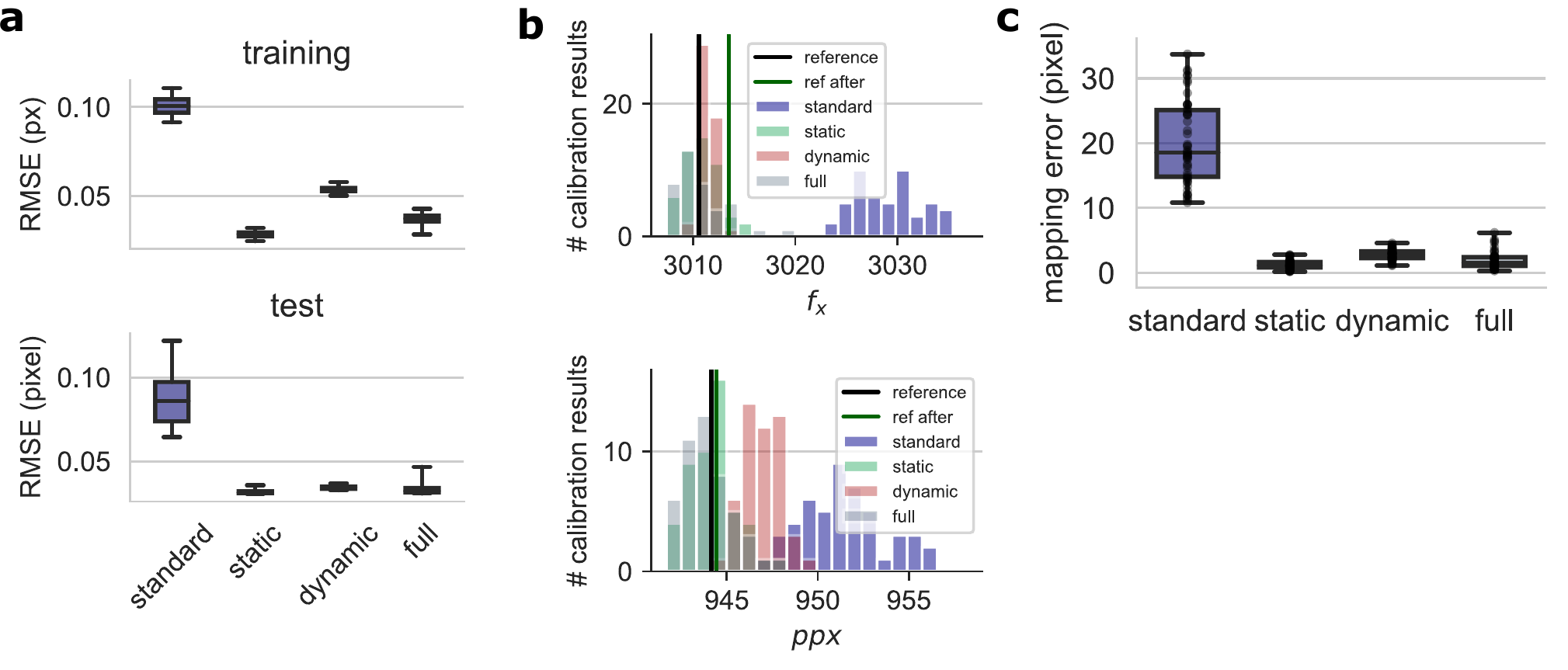}
	\caption{Calibration results on the reference dataset, where the target is lying statically on the ground while the camera is being moved. \textbf{a} RMSE on the calibration dataset (training) and the test dataset (test). \textbf{b} Distribution of the estimates of two exemplary parameters (focal length $f_x$ and principal point ppx). \textbf{c} Mapping error of calibration results w.r.t. reference calibration.}
	\label{S1}
\end{figure*}

\begin{table}[ht!]
	\centering
	\begin{tabular}{ lll } \toprule
		target & calibration method & loop closure RMSE (px) \\ \midrule
		reference &  standard & $12.4 \pm 2.5 $  \\ 
		reference &  static & $1.7 \pm 0.6$  \\ 
		reference  & dynamic (ours) & $1.0 \pm 0.4$  \\ 
		reference  & full (ours) & $2.5 \pm 1.3$ \\ 
		
		\bottomrule
	\end{tabular}
	\vspace{2mm}
	\caption{Reprojection error (root mean squared error, RMSE) in the Structure-from-Motion experiment. The table corresponds to table~1 in the main paper, and shows the corresponding results on the reference dataset where the target is lying statically on the ground. The error reflects the loop closure error of the trajectory in image space. The values are mean and standard deviation across 50 calibrations. }
\end{table}

\begin{table}[ht!]
	\centering
	\begin{tabular}{ ll } \toprule
		target &  maximum out-of-plane deformation (m) \\ \midrule
		T1  &  $0.0026 \pm 0.0012$\\ 
		T2 &   $0.0017 \pm 0.0024 $  \\ 
		T3  &  $0.0030 \pm 0.0021$  \\ 
		T4  &  $0.00019 \pm 0.00057$ \\ 
		
		\bottomrule
	\end{tabular}
	\vspace{2mm}
	\caption{Maximum estimated out-of-plane deformation for the different targets, using the \emph{dynamic} method. The values are mean and standard deviation of the maximum deformation $\Delta z$ in each image, across all 50 subsets for each target. The small, light target T4 showed the smallest deformations, while the larger targets (T1 and T2), as well as the self-made target T3 showed larger deviations.}
	\label{table:2}
\end{table}

\begin{figure*}[h!]
	\centering
	\includegraphics[width=0.99\textwidth]{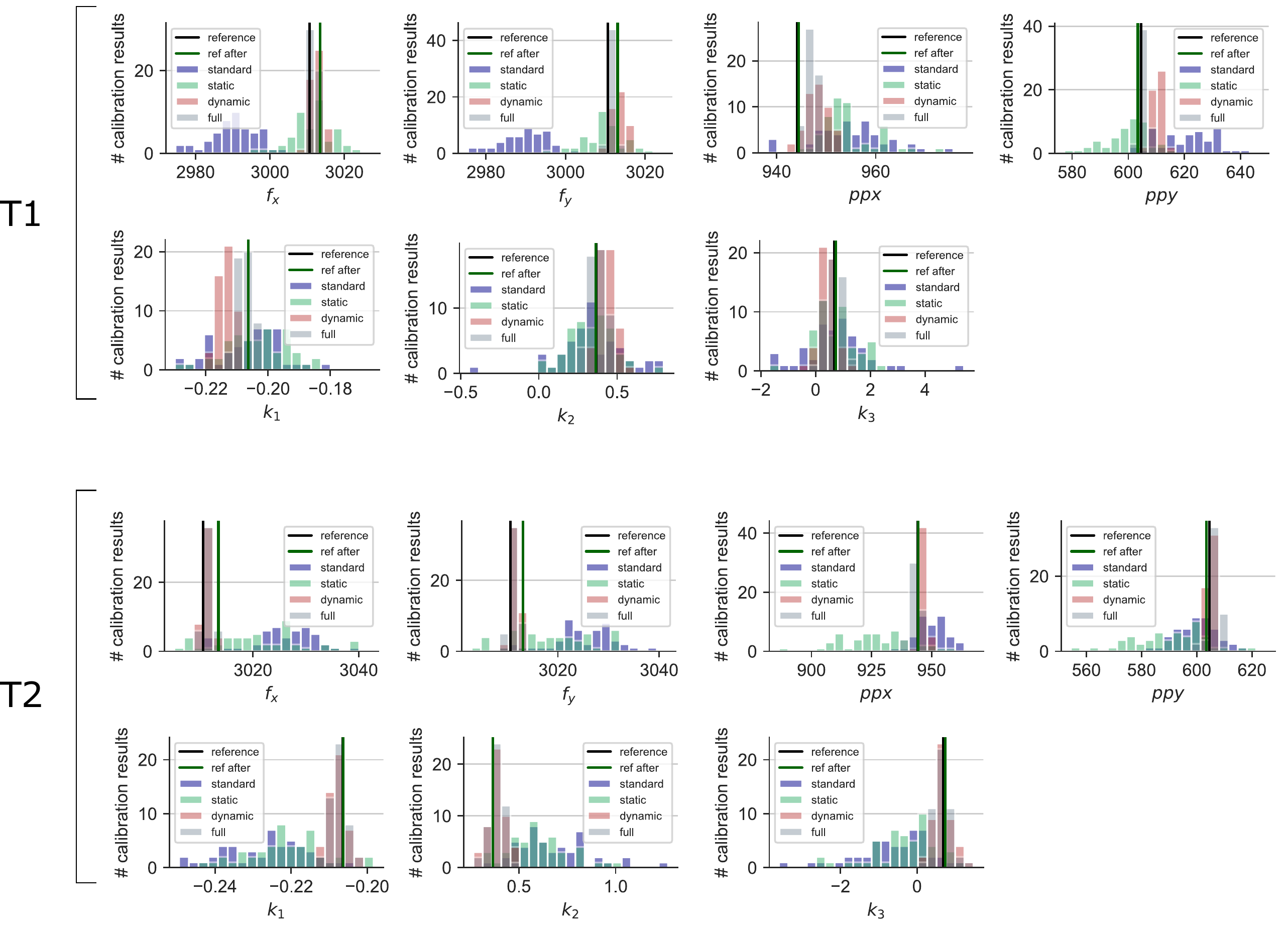}
	\caption{Distribution of estimated intrinsic parameters for targets T1 and T2, using the four different methods \emph{standard}, \emph{static}, \emph{dynamic} and \emph{full}.}
	\label{S2a}
\end{figure*}

\begin{figure*}[h!]
	\centering
	\includegraphics[width=0.99\textwidth]{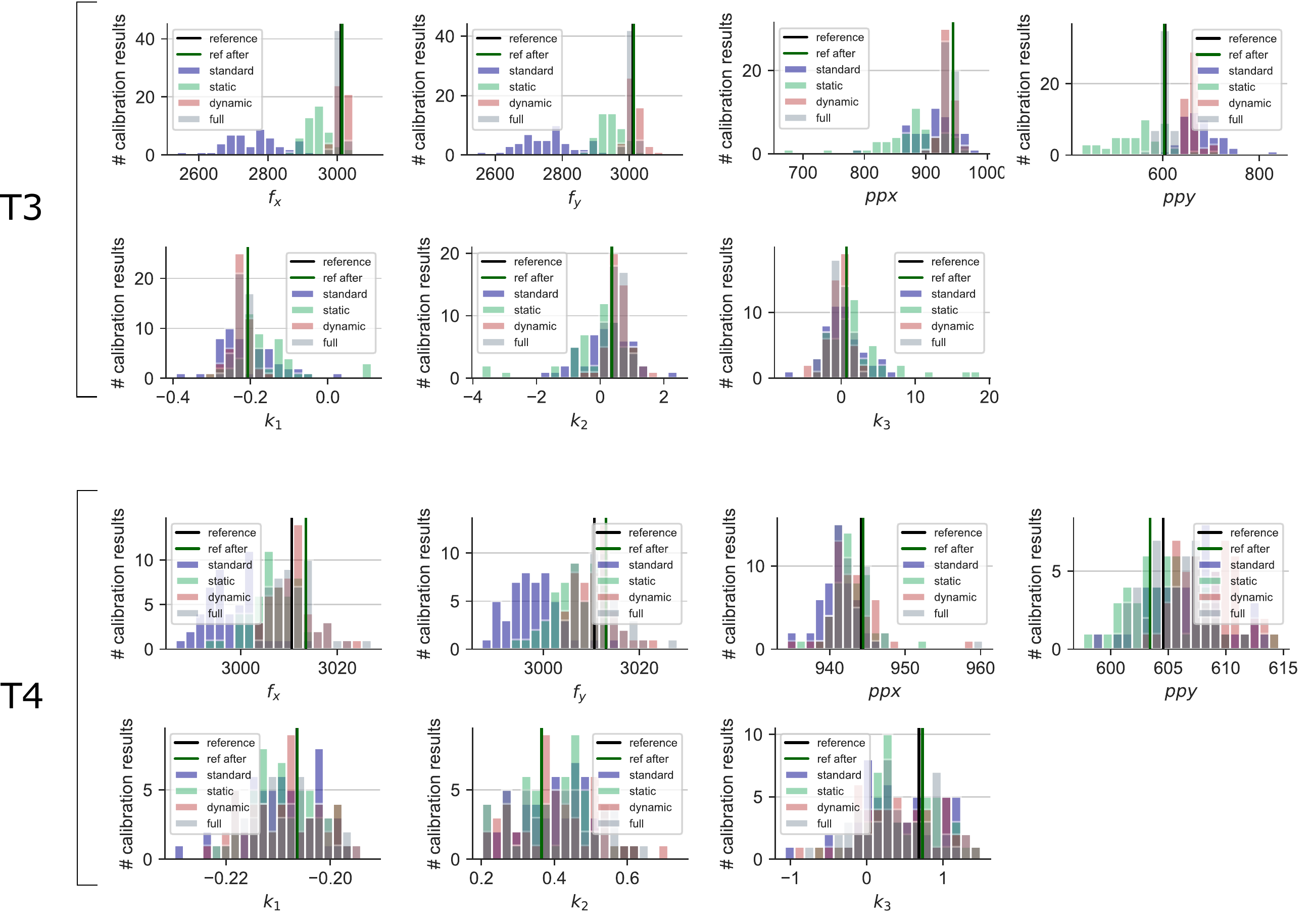}
	\caption{Distribution of estimated intrinsic parameters for targets T3 and T4, using the four different methods \emph{standard}, \emph{static}, \emph{dynamic} and \emph{full}.}
	\label{S2b}
\end{figure*}

\end{document}